\documentclass[conference]{IEEEtran}
\IEEEoverridecommandlockouts

\usepackage{cite}
\usepackage{amsmath,amssymb,amsfonts}
\usepackage{algorithmic}
\usepackage{graphicx}
\usepackage{textcomp}
\usepackage{xcolor}
\usepackage{booktabs}
\usepackage{subcaption}
\usepackage[acronym]{glossaries}
\glsdisablehyper
\newacronym{sota}{SOTA}{state-of-the-art}
\newacronym{kpi}{KPI}{key performance indicator}

\newacronym{pac}{PAC}{probably approximately correct}
\newacronym{rv}{RV}{random variable}
\newacronym{iid}{i.i.d.}{independent and identically distributed}

\newacronym{fpr}{FPR}{false positive rate}
\newacronym{tpr}{TPR}{true positive rate}
\newacronym{roc}{ROC}{receiver operating characteristic}
\newacronym{auc}{AUC}{area under curve}
\newacronym{ece}{ECE}{expected calibration error}
\newacronym{kl}{KL}{Kullback-Leibler}
\newacronym{rms}{RMS}{root mean-square}
\newacronym{mav}{MAV}{mean-absolute value}

\newacronym{lti}{LTI}{linear time-invariant}

\newacronym{ai}{AI}{artificial intelligence}
\newacronym{nn}{NN}{neural network}
\newacronym{rnn}{RNN}{recurrent neural network}
\newacronym{ml}{ML}{maximum likelihood}
\newacronym{map}{MAP}{maximum a posteriori}
\newacronym{ood}{OoD}{out-of-distribution}
\newacronym{gd}{GD}{gradient descent}
\newacronym{erm}{ERM}{empirical risk minimization}
\newacronym{mlp}{MLP}{multi-layer perceptron}
\newacronym{vae}{VAE}{variational auto-encoder}
\newacronym{cdl}{CDL}{contrastive density learning}
\newacronym{elbo}{ELBO}{evidence lower bound}
\newacronym{vi}{VI}{variational inference}
\newacronym{mcmc}{MCMC}{Markov chain Monte Carlo}
\newacronym{em}{EM}{expectation maximization}
\newacronym{vem}{VEM}{variational expectation maximization}
\newacronym{lvm}{LVM}{latent variable model}
\newacronym{hmm}{HMM}{hidden Markov model}
\newacronym{gp}{GP}{Gaussian process}
\newacronym{bnn}{BNN}{Bayesian neural network}
\newacronym{fwp}{FWP}{fast weight programmer}
\newacronym{eb}{EB}{empirical Bayes}
\newacronym{kd}{KD}{knowledge distillation}

\newacronym{rkhs}{RKHS}{reproducible kernel Hilbert space}

\newacronym{ar}{AR}{autoregressive}
\newacronym{llm}{LLM}{large language model}
\newacronym{icl}{ICL}{in-context learning}
\newacronym{la}{LA}{linear attention}
\newacronym{ssm}{SSM}{state-space model}
\newacronym{s4}{S4}{structured state-space sequence}
\newacronym{fla}{FLA}{flash linear attention}
\newacronym{gla}{GLA}{gated linear attention}
\newacronym{dyt}{DyT}{dynamic $\mathrm{tanh}$}

\newacronym{ptq}{PTQ}{post-training quantization}
\newacronym{qat}{QAT}{quantization-aware training}
\newacronym{lsq}{LSQ}{learned step-size quantization}

\newacronym{snn}{SNN}{spiking neural network}
\newacronym{ann}{ANN}{artificial neural network}
\newacronym{if}{IF}{integrate-and-fire}
\newacronym{lif}{LIF}{leaky integrate-and-fire}
\newacronym{a2s}{A2S}{ANN-to-SNN}
\newacronym{sop}{SOP}{synaptic operation}

\newacronym{gpu}{GPU}{graphics processing unit}
\newacronym{hbm}{HBM}{high bandwidth memory}
\newacronym{flop}{FLOP}{floating-point operation}
\newacronym{lut}{LUT}{look-up table}
\newacronym{acc}{ACC}{accumulation}
\newacronym{bbm}{BBM}{binary-binary multiplication}
\newacronym{qbsm}{QBSM}{quantized-binary static multiplication}
\newacronym{qsm}{QSM}{quantized static multiplication}
\newacronym{qdm}{QDM}{quantized dynamic multiplication}
\newacronym{qnl}{QNL}{quantized non-linearity}
\newacronym{fp}{FP}{floating-point}
\newacronym{fixp}{FixP}{fixed-point}

\newacronym{mdlm}{MDLM}{masked diffusion language model}
\newacronym{nmdlm}{N-MDLM}{neuromorphic masked diffusion language model}
\newacronym{e2d2}{E2D2}{efficient encoder-decoder diffusion}
\newacronym{icms}{ICMS}{in-chip memory system}
\newacronym{ocms}{OCMS}{off-chip memory system}
\newcommand{\glsemph}[1]{%
  \glsunset{#1}%
  \emph{\glsentrylong{#1}} (\glsentryshort{#1})%
}
\newcommand{\glsemphpl}[1]{%
  \glsunset{#1}%
  \emph{\glsentrylongpl{#1}} (\glsentryshortpl{#1})%
}

\newcommand{\R}{\mathbb{R}}
\newcommand{\dset}[1]{\{1, ..., #1\}}

\newcommand{\data}{\mathrm{data}}
\newcommand{\ops}{\mathrm{ops}}
\newcommand{\enc}{\mathrm{enc}}
\newcommand{\dec}{\mathrm{dec}}
\newcommand{\emb}{\mathrm{emb}}
\newcommand{\hid}{\mathrm{hid}}
\newcommand{\TT}{\mathrm{TT}}
\newcommand{\TL}{\mathrm{TL}}

\newcommand{\ee}{\mathrm{e2e}}

\begin{document}
\bstctlcite{IEEEexample:BSTcontrol} 

\title{Neuromorphic Diffusion Language Models:  Addressing Compute and Memory Bottlenecks via Sparsity and Block Denoising}

\author{Dengyu Wu$^{1}$, Clement Ruah$^{2}$, Jiechen Chen$^{1}$, Bipin Rajendran$^{2}$,  and Osvaldo Simeone$^{2}$ \\
${}^{1}$Department of Engineering, King’s College London, London, UK \\
${}^{2}$Institute for Intelligent Networked Systems (INSI), Northeastern University London, London, UK \\
E-mails:\{dengyu.wu, jiechen.chen\}@kcl.ac.uk, \{c.ruah, b.rajendran, o.simeone\}@nulondon.ac.uk
\thanks{This work was supported by the European Research Council (ERC) under the European Union’s Horizon Europe Programme (grant agreement No. 101198347), by Open Fellowships of the EPSRC (EP/W024101/1,  EP/X011356/1), by the EPSRC project (EP/X011852/1).
}
\vspace*{-0.35cm}
}

\maketitle

\begin{abstract}
\Gls{ar} \glspl{llm} are inherently inefficient at inference time because each generated token requires accessing the full set of model parameters, leading to low operational intensity and high energy consumption. Masked diffusion language models (MDLMs) partially address this limitation for memory-bound settings by allowing multiple tokens to be generated per parameter access. In order to further enhance inference efficiency on modern platforms with extensive in-chip memory, this work proposes  neuromorphic MDLMs (N-MDLMs), which integrate block diffusion with spike-based neuromorphic computation to jointly improve throughput and energy efficiency.  While block diffusion increases token throughput by producing multiple tokens per parameter access, spike-induced sparsity reduces effective parameter traffic and computations by skipping inactive channels. To analyze the synergistic effect of sparsity and diffusion, we develop a token-level roofline-inspired model that captures the combined impact of block-parallel generation and spike sparsity on decoding efficiency. Experimental results on translation tasks show that, thanks to spike-induced sparsity,  N-MDLMs achieve substantial improvements in energy efficiency and throughput even in compute-bound platforms  for which MDLMs would fail to improve over AR-LLMs. 
\glsreset{ar}\glsreset{llm}\glsreset{mdlm}\glsreset{nmdlm}
\end{abstract}

\begin{IEEEkeywords}
neuromorphic diffusion model, spiking neural network, neuromorphic LLM
\end{IEEEkeywords}

\section{Introduction}
\label{sec:intro}

\Glspl{llm} are increasingly deployed in latency-sensitive and cost-constrained settings, where inference efficiency, in terms of memory utilization, throughput, and energy consumption, determines system performance. However, conventional \glsemph{ar} \glspl{llm} require repeatedly accessing the full set of model parameters, resulting in limited operational intensity and low throughput ~\cite{bi2026rooflinebench,verhelst2025keep}.

To alleviate these limitations, \glsemphpl{mdlm} generate multiple tokens per parameter access \cite{arriola2025block}. However, energy efficiency is still constrained by the use of transformer models in each denoising step. Furthermore, throughput benefits hinge on the system operating in a memory-bound regime, and may not translate to modern inference platforms with substantial in-chip memory \cite{appuswamy2024breakthrough, abts2022software}.

\begin{figure}
    \centering
    \includegraphics[width=0.9\linewidth]{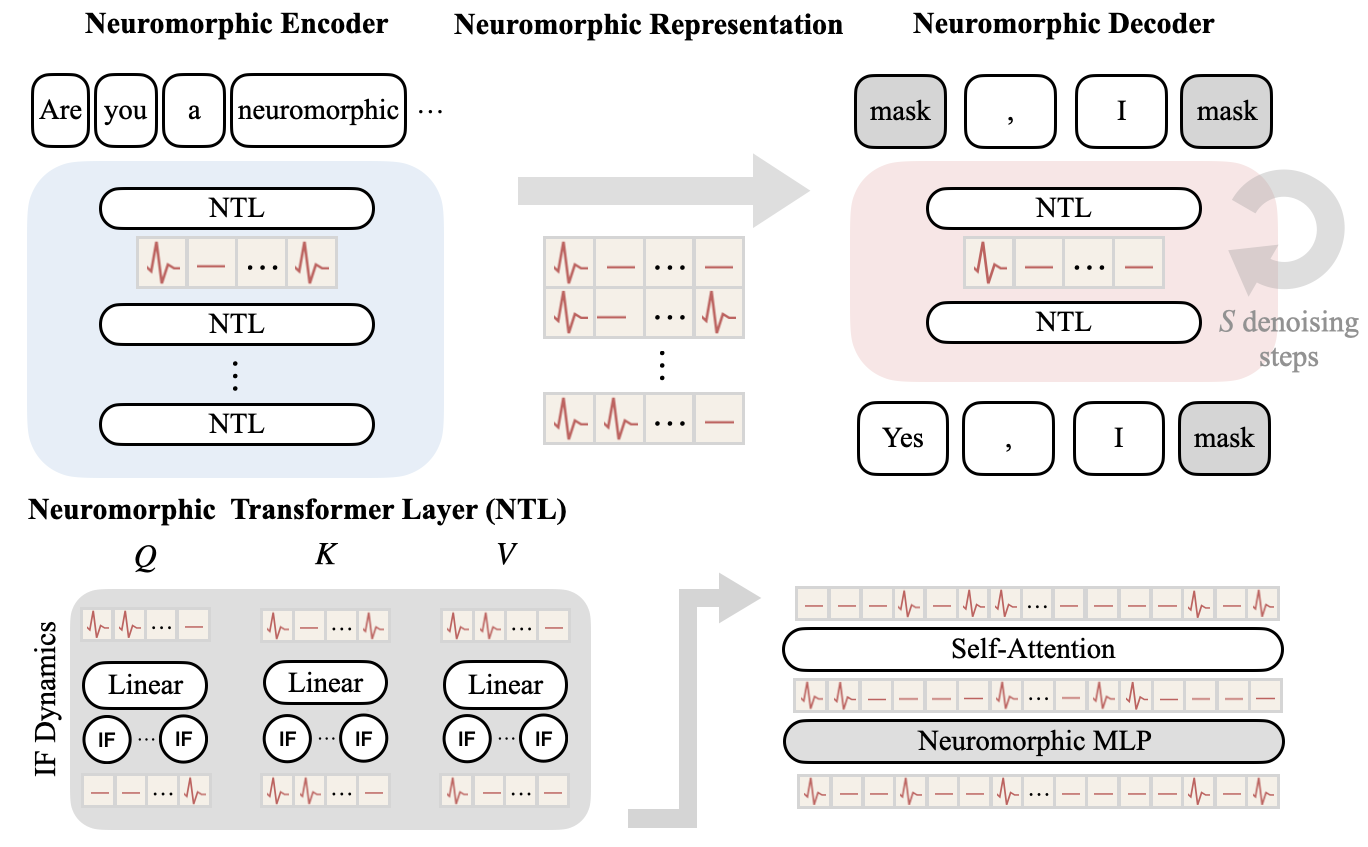}
    \caption{In the proposed neuromorphic masked diffusion language model (N-MDLM), a neuromorphic encoder processes the clean prefix tokens and produces neuromorphic contextual representations.
    The neuromorphic decoder iteratively refines a block of masked tokens through denoising steps while reusing the neuromorphic representations from the encoder. Each neuromorphic transformer layer is implemented with IF neuron dynamics (MLP = multilayer perceptron).}
    \label{fig:intro_spike_arch}
    \vspace{-10pt}
\end{figure}

\emph{Neuromorphic computing} principles have recently been shown to offer potential benefits in reducing energy consumption for standard \gls{ar} transformers \cite{zhou2022spikformer,simeone2026modern,pan2025spikingbrain}. Neuromorphic models operate in an event-driven manner, with computation triggered by events, also known as spikes, representing discrete activations. In the absence of a spike, tailored hardware deployments can avoid computation and memory access, benefiting from dynamic sparsity \cite{simeone2026modern}.

\begin{figure}[!t]
    \centering
    \includegraphics[width=0.85\linewidth]{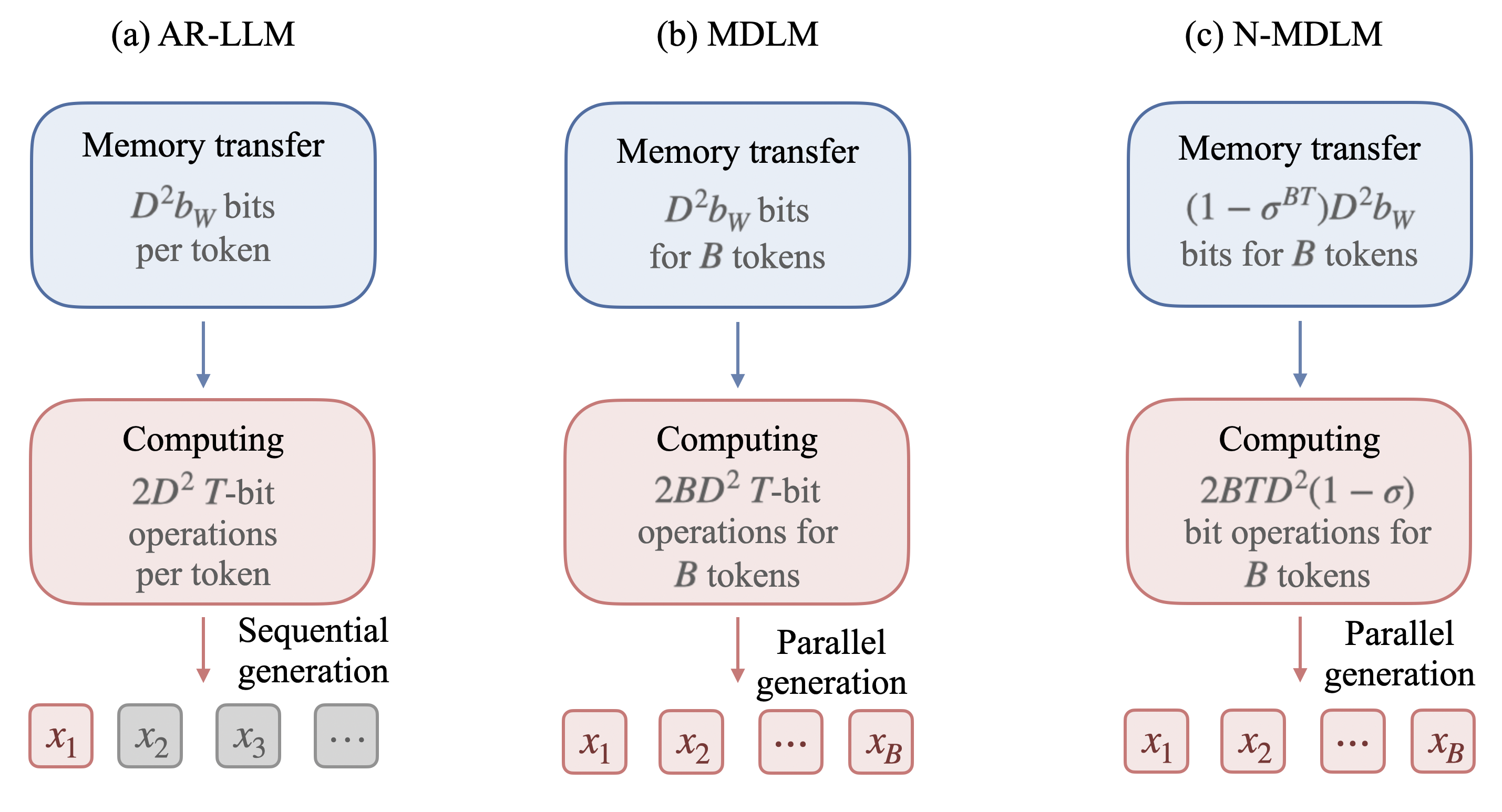}
    \caption{From \gls{ar}-\gls{llm} to the proposed \gls{nmdlm}: 
    (a) An \gls{ar}-\gls{llm} requires accessing the full set of model parameters for each generated token. 
    (b) An \gls{mdlm} generates $B$ tokens per parameter access, increasing the number of operations per access proportionally to $B$. 
    (c) The proposed \gls{nmdlm} further exploits spiking sparsity, such that only active channels trigger synaptic operations, thereby reducing both parameter traffic and computational cost.}
    \label{fig:intro}
    \vspace{-10pt}
\end{figure}

In this work, as shown in Fig.~\ref{fig:intro_spike_arch}, we introduce \glsemphpl{nmdlm}, a model class that enhances throughput via parallel denoising steps, while further improving inference efficiency via the dynamic sparsity afforded by neuromorphic computing. As illustrated in Fig.~\ref{fig:intro}, while conventional \gls{ar}-\glspl{llm} generate one token per parameter access, \glspl{mdlm} amortize this cost across multiple tokens within a block \cite{arriola2025block}. \Glspl{nmdlm} further enhance operational intensity through event-driven sparsity, thereby reducing both parameter traffic and computational cost. Even on modern neuromorphic hardware \cite{davies2018loihi, appuswamy2024breakthrough} and \gls{llm} inference accelerators \cite{abts2022software} that co-locate memory and compute inside the chip,  this can shift the operation towards a memory-bound regime that makes parallel generation via diffusion both throughput and energy efficient.

The main contributions of this work are as follows:
\begin{itemize}
    \item \emph{Neuromorphic \gls{mdlm}:} We introduce \glspl{nmdlm}, a novel class of neuromorphic \glspl{llm} obtained via conversion of a conventional \gls{mdlm} \cite{arriola2025block} based on quantization and \gls{if} spiking models \cite{pan2025spikingbrain}.
    \item \emph{Analysis:} We develop a token-level roofline-inspired analytical framework that characterizes the achievable decoding efficiency of N-MDLMs as a function of block size and activation sparsity, capturing the joint and synergistic effects of block-parallel generation and event-driven computation on both computation and memory access.
    \item \emph{Experimental validation:} We validate the proposed framework on translation tasks, demonstrating that N-MDLMs  achieve improved token throughput and energy efficiency compared to AR-LLMs and conventional MDLMs, while maintaining competitive task performance. The gains are seen to arise from a synergistic use of sparsity, to alleviate bottlenecks due to  computational limitations, and diffusion, to mitigate throughput constraints due to memory transfer.
\end{itemize}

\section{Preliminaries}
In this section, we start by reviewing AR-LLMs and MDLMs, as well as the IF neuron model.
\subsection{Autoregressive vs. Masked Diffusion Language Models}
\label{sec:background}

\gls{ar}-\glspl{llm} factorize the joint distribution $p_\theta(x)$ over the token sequence $x=(x_1,\ldots,x_L)$ as
\begin{equation}\label{eq:ar}
p_\theta(x) = \prod_{l=1}^{L} p_\theta(x_l \mid x_{<l}),
\end{equation}
where $\theta$ denotes the model parameters and each token $x_l$ takes values in the vocabulary $\mathcal{V}$.
In \eqref{eq:ar}, the conditional distribution $p_{\theta}(x_l | x_{<l})$ assigns a probability to token $x_l$ given the prefix $x_{<l}=(x_1,\ldots,x_{l-1})$ consisting of previous tokens.
The factorization in \eqref{eq:ar} induces a strictly sequential inference procedure such that at each decoding step the model produces a
single token $x_l$.
Accordingly, as illustrated in Fig.~\ref{fig:intro}(a), each parameter access yields one output token, limiting operational intensity and throughput.

\Glspl{mdlm}~\cite{arriola2025block} relax the sequential operation by grouping tokens into blocks of size $B$ tokens and generating each block in parallel via a number of denoising steps (see Section \ref{sec:architecture}).
Specifically, the sequence $x=(x_1,\ldots,x_L)$ is partitioned into blocks $x^b=(x_{b,1},\ldots,x_{b,B})$ with $b=1, \ldots, \lceil L/B \rceil$, and the joint distribution is factorized at the block level as
\begin{equation}
p_\theta(x) =
\prod_{b=1}^{\lceil L/B \rceil}
p_\theta(x^b \mid x^{<b}), \label{fac}
\end{equation}
where $x^{<b}=(x^1,\ldots,x^{b-1})$ denotes the previously generated blocks. 

\subsection{Neuromorphic Computing Principles}

Neuromorphic computing systems represent information using discrete spike events generated by neuron dynamics.
A commonly used neuron model is the \glsemph{if} neuron with a soft-reset mechanism \cite{wu2025optimizing}.
For any neuron, a spike event, corresponding to output $s_{t}=1$, is generated when the internal state variable $V_{t}$ crosses a threshold $V_{\mathrm{th}}$ as
\begin{equation}
s_{t} =
\begin{cases}
1, & V_{t} \ge V_{\mathrm{th}}, \\
0, & V_{t} < V_{\mathrm{th}} .
\end{cases}
\label{spike}
\end{equation} 
Given an input synaptic current $I_{t}$, the state variable evolves as
\begin{equation}
V_{t+1} = V_{t} + I_{t} - s_{t} V_{\mathrm{th}}.
\label{if}
\end{equation}
The synaptic current $I_{t}$ results from incoming spikes from presynaptic neurons interacting with their synaptic weights. 

Under event-driven execution, synaptic operations are triggered only when spikes occur.
Consequently, inactive neurons incur neither computation nor memory access, leading to \emph{dynamic sparsity} in both compute and parameter traffic \cite{davies2018loihi}.

\section{Neuromorphic Masked Diffusion Models}

In this section, we introduce \glspl{nmdlm}, a neuromorphic block diffusion \gls{llm} constructed by converting existing conventional encoder-decoder-based \glspl{mdlm} \cite{pan2025spikingbrain}.

\subsection{Architecture} \label{sec:architecture}

As illustrated in Fig. \ref{fig:intro_spike_arch}, the proposed \gls{nmdlm} follows a block diffusion generation paradigm in which multiple tokens are predicted within a block through an iterative denoising process. We specifically adopt the encoder-decoder architecture from \cite{arriola2025encoderdecoder}.
In it, the encoder processes the current context to produce a representation that serves as conditioning information for the decoder during the denoising process.

At encoding stage $b$ based on factorization \eqref{fac}, the model generates block $x^b$ given context $x^{<b}=(x^1,\ldots,x^{b-1})$.
The encoder $f_{\theta_{\mathrm{enc}}}(\cdot)$ first processes the context $x^{<b}$ to produce the latent representation
\begin{equation}
h^b = f_{\theta_{\mathrm{enc}}}(x^{<b}).
\end{equation}

Using the latent representation $h^b$, generation at the decoder starts from initially masked tokens representation $z_0^b \in \mathbb{R}^{B \times D}$, where $D$ denotes the hidden dimension of the transformer.
The matrix $z_0^b$ contains by column the embedding of the special mask token for all $B$ tokens in the block.
This initial uninformative signal is iteratively refined over $S$ denoising steps indexed as $s=1, \ldots, S$.
At each denoising step $s$, $B/S$ tokens, with ratio $B/S$ assumed to be an integer, are selected for remaining masked position to be unmasked \cite{arriola2025block}.
To this end, the transformer decoder $f_{\theta_{\mathrm{dec}}}(\cdot)$ maps the contextual representation $h^{b}$ and the current noisy block $z_s^b=(z_{s,1}^b,\ldots,z_{s,B}^b)$ to produce updated token embeddings
\begin{equation}
z_{s+1}^b = f_{\theta_{\mathrm{dec}}}(z_s^b,\; h^b) \in  \mathbb{R}^{B \times D},
\end{equation}
with $B/S$ additional tokens unmasked as compared to  $z_{s}^b$.

\subsection{Conversion} \label{sec:conversion}

Starting from a pre-trained encoder-decoder \gls{mdlm}~\cite{pan2025spikingbrain}, we obtain an N-MDLM via a quantization-based conversion methodology that leverages the \gls{if} neuron model \eqref{if} \cite{pan2025spikingbrain}.
The main principle underlying this approach is that the spiking rate of an \gls{if} neuron, or variants thereof, can match the output of a quantization function in which the quantization step size equals the threshold $V_{\rm th}$ in \eqref{spike} \cite{pan2025spikingbrain}.

Given an activation vector $\boldsymbol{a}=(a_1,\ldots,a_D)$ produced by a layer of the encoder or decoder, a quantization function with an adaptive step size obtains a quantized integer-valued vector $\hat{\boldsymbol{a}}\in \mathbb{Z}^D$ as \cite{pan2025spikingbrain}
\begin{equation}\label{eq:adaptive_threshold}
\hat{\boldsymbol{a}} =
\mathrm{round}\!\left(
\frac{\boldsymbol{a}}{V_{\mathrm{th}}(\boldsymbol{a})}
\right),
\qquad
V_{\mathrm{th}}(\boldsymbol{a}) =
\frac{1}{K}\,\mathrm{mean}(|\boldsymbol{a}|), 
\end{equation}
where the rounding function $\mathrm{round}(\cdot)$ and the magnitude function $|\cdot|$ are applied element-wise.
The parameter $K$ in \eqref{eq:adaptive_threshold} controls the number of quantization levels, with larger values of $K$ increasing the resolution.

Each quantized value $\hat{a}_i$ is represented via bitwise coding, a temporal coding scheme using ternary spikes $s_{i,t} \in \{-1,0,1\}$ over $T$ time steps.
The spikes are related to integer $\hat{a}_i$ as
\begin{equation}
\label{eq:activation_spiking}
\hat{a}_i = \sum_{t=1}^{T} 2^{t-1}\, s_{i,t},
\end{equation}
where $T = \left\lfloor \log_2 \left(\max_i |\hat{a}_i|\right) \right\rfloor$ denotes the number of time steps. 

Let $\boldsymbol{w} =[w_1,...,w_D]^\top\in \mathbb{R}^{D}$ be a weight vector.
This may correspond to the row of a linear layer or to the linear projection in an MLP. The inner product $\boldsymbol{w}^\top\hat{\boldsymbol{a}}$ for a quantized activation vector $\hat{\boldsymbol{a}}$ can be evaluated using a modified IF neuron operating on the spiking signals in \eqref{eq:activation_spiking}.
To this end, let $I_{t}= \sum_{i=1}^D w_{i} s_{i,t}$ be the input current to the IF neuron at timestep $t$, and consider the modified IF neuron with dynamics
\begin{equation}
V_{t+1} = V_{t} + 2^{t} I_{t} - 2^{t} \hat{s}_{t} V_{\mathrm{th}}(\boldsymbol{a}),
\end{equation}
where $\hat{s}_{t}$ denotes the emitted spike, computed as
\begin{equation}
\label{eq:spike}
\hat s_{t} =
\begin{cases}
1, & V_{t} \ge 2^t V_{\mathrm{th}}(\boldsymbol{a}), \\
-1, & V_{t} \le -2^t V_{\mathrm{th}}(\boldsymbol{a}), \\
0, & \text{otherwise}.
\end{cases}
\end{equation}
As a result, the linear transformation can be approximately  reconstructed as \cite{pan2025spikingbrain}
\begin{equation} \label{eq:linear}
\boldsymbol{w}^\top \hat{\boldsymbol{a}} \approx
V_{\mathrm{th}}(\boldsymbol{a}) \sum_{t=1}^{T} 2^{t-1}\, \hat{s}_{i,t}.
\end{equation}

\section{Performance Analysis} \label{sec:roofline}

Compared to conventional \gls{ar}-\glspl{llm}, \glspl{nmdlm} can modulate data access and computation via the choice of block size $B$, amortizing the parameter accesses per block, and the quantization hyperparameter $K$, controlling the sparsity of generated spikes.
In this section, we study the effect of block size $B$ and sparsity hyperparameter $K$ on the per-token number of operations and memory transfer requirements, yielding an estimated token throughput, within a roofline-inspired framework \cite{verhelst2025keep}.

\subsection{Computational and Memory-Transfer Load}

Following \cite{bi2026rooflinebench}, we assume that the cost of a forward pass in all models is dominated by operations in linear layers.
For parallel computation of $B$ tokens, these are represented by operations of the form $Y = X W^\top \in \R^{BT \times D}$, with parameter matrix $W \in \R^{D \times D}$ and input $X \in \R^{B T \times D}$ obtained by concatenating the spike train encoding $B$ consecutive tokens over $T$ time steps.
We formally define the sparsity level as the probability $\sigma = P(x_{t, d} = 0) \in [0, 1]$ that an element $x_{t, d}$ from matrix $X$ is zero, assuming all entries $x_{t, d}$ to be independent and identically distributed for all $t \in \dset{BT}$ and $d \in \dset{D}$.
The sparsity level $\sigma$ is controlled by the hyperparameter $K$, with a larger $K$ generally reducing $\sigma$ \cite{pan2025spikingbrain}.

We further make the following event-driven computation assumptions \cite{davies2018loihi}: $(i)$ only spike events $x_{t, d} = 1$ contribute to the computation load; and $(ii)$ the $d$-th row of matrix $W$ is loaded only if there is at least one spike $x_{t, d} = 1$ within the corresponding batch channel sequence $X_d = \{ x_{t, d} \}_{t \in \dset{BT}}$.

\subsubsection{Per-Token Computational Load}

Assumption $(i)$ implies that each element in the output $Y$ needs on average $2 D (1 - \sigma)$ operations, counting additions and multiplications separately \cite{bi2026rooflinebench}.
Accordingly, the total average number of operations per token is given by
\begin{equation}
\label{eq:n_ops}
    N_{\ops}(B, \sigma) = 2 T D^2 (1 - \sigma).
\end{equation}

\subsubsection{Per-Token Memory-Transfer Load}

Assumption $(ii)$ implies that each row of matrix $W$ is loaded with probability $1 - \sigma^{BT}$.
Accordingly, the average number of bits to be loaded from memory per token is
\begin{equation}
\label{eq:n_data}
    N_{\data}(B, \sigma) = \frac{D^2 (1 - \sigma^{BT})}{B} b_W,
\end{equation}
where $b_W \geq 1$ is the number of bits per weight.
Note that we have ignored the cost of transferring $X$ and $Y$ from and to memory as, in practice, it is negligible compared to the cost of loading the weights \cite{bi2026rooflinebench}. The quantities in \eqref{eq:n_ops} and \eqref{eq:n_data} also apply to linear layers in \gls{ar}-\glspl{llm}, with $B=1$ and $\sigma = 0$, and \glspl{mdlm}, with $B \geq 1$ and $\sigma = 0$.

\subsection{Token Throughput}

For a hardware platform capable of producing $L_\ops$ operations per second and of moving $L_\data$ bits per second, the theoretical attainable token latency (TL) for a \gls{nmdlm} linear layer is
\begin{equation}
\label{eq:latency}
\TL(B, \sigma) = \max \bigg\{
    \underbrace{\frac{N_{\ops}(B, \sigma)}{L_\ops}}_{\text{compute-bound}},
    \underbrace{\frac{N_{\data}(B, \sigma)}{L_\data}}_{\text{memory-bound}}
\bigg\},
\end{equation}
which is measured in seconds per token.
The end-to-end latency of the model can then be approximated by summing the individual latencies $\TL(B, \sigma)$ of each linear layer during a forward pass.
When the equality $\TL(B, \sigma) = N_{\ops}(B, \sigma) / L_\ops$ holds, the system is said to be in a \emph{compute-bound regime}, and throughput gains can only come from an increase in the computation speed $L_\ops$.
Conversely, when the equation $\TL(B, \sigma) = N_{\data}(B, \sigma) / (B L_\data)$ holds, the system is in a \emph{memory-bound regime}, and throughput can be improved by increasing the memory bandwidth $L_\data$.

The expression \eqref{eq:latency} highlights a \emph{regime-dependent} tradeoff between the block size $B$ and the sparsity level $\sigma$: when the system is compute-bound, the latency gains in \glspl{nmdlm} are linearly proportional to the sparsity level $\sigma$ regardless of the block size $B$. In contrast, when the system is in a memory-bound regime, increasing the block size $B$ reduces the benefits of sparsity, with the effective data-loading sparsity reduced to $\sigma^{BT} < \sigma$.
Moreover, increasing the block size $B$ steers the system towards a compute-bound regime,
while increasing the sparsity level $\sigma$ steers it towards a memory-bound regime. 
Therefore, in order to get maximal benefits from sparsity, \glspl{nmdlm} should ideally run on systems with high memory bandwidth $L_\data$, such as near-memory compute neuromorphic devices \cite{davies2018loihi, appuswamy2024breakthrough}, using a larger block size $B$ as a means to ease the system into compute-bound regimes.

\section{Experiments}
\label{sec:experiments}

In this section, we validate the performance of \glspl{nmdlm} via experimental results on the \textsc{WMT 14 de-en} translation task \cite{bojar-etal-2014-findings}. All experiments are conducted on an NVIDIA DGX Spark \gls{gpu}.

\subsection{Setup}

For translation experiments on the \textsc{WMT 14 de-en} dataset \cite{bojar-etal-2014-findings} we adopt the \gls{e2d2} architecture in \cite{arriola2025encoderdecoder} with $N_\enc = 28$ encoder layers and $N_\dec = 4$ decoder layers, embedding dimension $D_\emb = 512$ and hidden intermediate dimension $D_\hid = 1536$ (approximately 250M parameters) \cite{arriola2025encoderdecoder}.
The \gls{nmdlm} introduced in Sec.~\ref{sec:architecture} is simulated on \gls{gpu} using bitwise coding for $T = 8$ time steps.
After applying the  conversion pipeline described in Section \ref{sec:conversion} starting from the pretrained model released by \cite{arriola2025encoderdecoder}, we perform an additional round of fine-tuning. 
Fine-tuning uses the Adam optimizer over $1000$ steps, with batch size $16$ and learning rate $3\times10^{-5}$ on the \textsc{WMT 14 de-en} dataset. In the experiments, we consider the setting $S=B$, so that one token is unmasked at each denoising step.
\vspace{-10pt}

\subsection{Performance Metrics}\label{sec:performance_metric}
We evaluate the performance of all models in terms of token throughput (token/s), energy per token (J/token), and accuracy. For throughput and energy, we leverage the analysis in Sec. \ref{sec:roofline}, together with sparsity estimates obtained by running the model on the mentioned \gls{gpu} platform.

\subsubsection{Sparsity, Per-Token Computational Load, and Per-Token Memory-Transfer Load} In order to apply the analysis in Sec. \ref{sec:roofline}, we empirically measure the spike sparsity $\sigma$ at each linear layer operation during inference.
In turn, the collected sparsity values are used to estimate theoretical number of spike operations $N_\ops(B, \sigma)$ (OPs) in \eqref{eq:n_ops} and data movement $N_\data(B, \sigma)$ (bits) in \eqref{eq:n_data}. We set the precision of weights to $b_W = 16$ bits.

To evaluate the end-to-end performance of the model, we denote as $N^{\enc(n, m)}_\ops$ and $N^{\enc(n, m)}_\data$  the respective number of estimated operations and memory transfers at the $m$-th linear layer of the $n$-th encoding layer, for $n \in \dset{N_\enc}$ and $m \in \dset{M_\enc(n)}$, where $M_\enc(n)$ denotes the number of linear layers at encoding layer $n$.
Similarly, we denote as $N^{\dec(n, m, k)}_\ops$ and $N^{\dec(n, m, k)}_\data$ the values estimated at the decoding layers for $n \in \dset{N_\dec}$ and $m \in \dset{M_\dec(n)}$, where $k \in \dset{S}$ indexes the denoising step. Generally, for a fixed layer position $(n, m)$, the values $N^{\dec(n, m, k)}_\ops$ and $N^{\enc(n, m, k)}_\data$ can vary across denoising steps $S$, as each step is characterized by a different sparsity level. These quantities are used to evaluate token throughput and energy per token, as discussed next. 

\subsubsection{Token Throughput}

Using the estimated number of computation operations and data transfers, we derive the corresponding token latencies using \eqref{eq:latency} as $\TL^{\enc(n, m)} = \max\{ N^{\enc(n, m)}_\ops / L_\ops, N^{\enc(n, m)}_\data / L_\data\}$ and $\TL^{\dec(n, m, k)} = \max\{ N^{\dec(n, m, k)}_\ops / L_\ops, N^{\dec(n, m, k)}_\data / L_\data\}$ for each individual linear operation in the encoder and the decoder, respectively. Recall that these quantities depend on the computational throughput $L_\ops$ and memory bandwidth $L_\data$.
The estimated end-to-end token throughput of the system is then computed by summing the latencies of individual layers as \begin{multline} \label{eq:e2e_latency}
\TT^\ee = \bigg(
    \sum_{n=1}^{N_\enc} \sum_{m=1}^{M_\enc(n)} \TL^{\enc(n, m)} \\
    + \sum_{n=1}^{N_\dec} \sum_{m=1}^{M_\dec(n)} \sum_{k=1}^{S} \TL^{\dec(n, m, k)}
\bigg)^{-1},
\end{multline}
which is measured in tokens per second.

\subsubsection{Energy per Token}

Similar to token throughput, the energy per token is estimated from the number of operations and memory transfers at the linear layers.
Given the energy $E_\ops$ consumed per operation and the energy $E_\data$ required  per bit transferred from memory, the total energy consumed per token is estimated as \cite{verhelst2025keep}
\begin{multline}\label{eq:energy}
E^\ee = 
    \sum_{n=1}^{N_\enc} \sum_{m=1}^{M_\enc(n)} \left\{ N^{\enc(n, m)}_\ops E_\ops + N^{\enc(n, m)}_\data E_\data \right\} \\
    + \sum_{n=1}^{N_\dec} \sum_{m=1}^{M_\dec(n)} \sum_{k=1}^{S} \left\{ N^{\dec(n, m, k)}_\ops E_\ops + N^{\dec(n, m, k)}_\data E_\data \right\}.
\end{multline}

\subsubsection{In-chip Memory vs. Off-chip Memory Systems}

\begin{table}[!ht]
\centering
\renewcommand{\arraystretch}{1.3}
\setlength{\tabcolsep}{4pt}
\begin{tabular}{|l|c|c|c|c|}   
    \hline
    System & $L_\ops$ (TOP/s) & $L_\data$ (Tb/s) & $E_\ops$ (pJ/OP) & $E_\data$ (pJ/b) \\
    \hline
    OCMS & $300$ & $8$ & $0.05$ & $10$ \\
    ICMS & $20$ & $30$ & $0.05$ & $0.5$ \\
    \hline
\end{tabular}
\caption{
    Hardware parameters for representative off-chip memory system (OCMS) and in-chip memory system (ICMS).}
\label{tab:hw_params}
\end{table}

The benefits of MDLMs were demonstrated in \cite{arriola2025encoderdecoder} for conventional memory-bound platforms, like GPUs, that are characterized by an off-chip memory system (OCMS).
In this work, we first validate the benefits of MDLMs on OCMS by adopting the hardware parameters, dictating computation rate $L_\ops$, memory bandwidth $L_\data$, energy per operation  $E_\ops$, and energy per memory transfer $E_\data$, in Table~\ref{tab:hw_params}, which are obtained from GPU metrics for tensor cores and \gls{hbm} \cite{nvidia2020a100}. 

Using this validation as a benchmark, we are interested in evaluating throughput and energy performance levels on an \gls{icms}, which is aligned with neuromorphic hardware \cite{davies2018loihi, appuswamy2024breakthrough} and \gls{llm} inference accelerators \cite{abts2022software} that co-locate memory and compute inside the chip.
Accordingly, the parameters for the \gls{icms} in Table \ref{tab:hw_params}  correspond to a significantly smaller compute-to-bandwidth ratio, i.e.,  $L_\ops / L_\data < 1$ OP/b.
Energy consumption parameters in Table \ref{tab:hw_params} are inspired by the numbers of in-chip and off-chip memory access in \cite{murmann2020mixed}.

\subsection{Results}

\begin{figure}[!ht]
    \centering
        \includegraphics[width=0.7\linewidth]{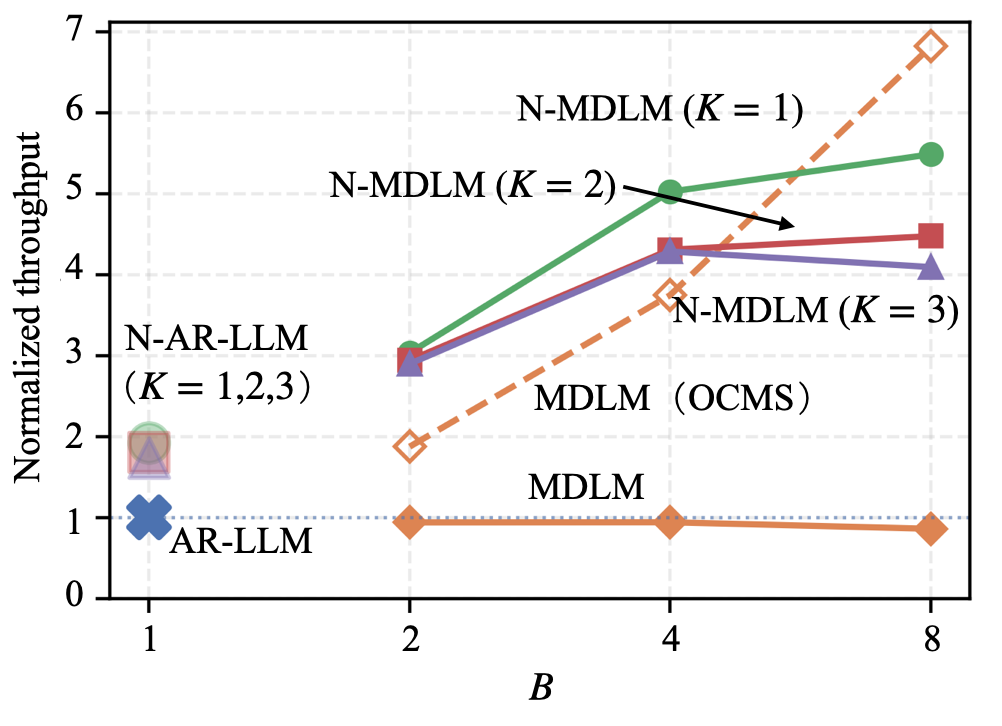}
        \caption{Normalized throughput versus block size $B$ for AR-LLM, MDLM, N-AR-LLM, and N-MDLM, when running on ICMS (solid lines) and OCMS (dashed lines). 
        Throughput, controlled by the sparsity parameter $K$ in \eqref{eq:adaptive_threshold}, is computed using \eqref{eq:e2e_latency} based on the computational and memory-transfer loads in \eqref{eq:n_ops} and \eqref{eq:n_data}. The throughput is normalized with respect to the AR-LLM baseline of the corresponding hardware.}
    \label{fig:wmt_throughputb}
\end{figure}

In Fig.~\ref{fig:wmt_throughputb}, we plot the end-to-end throughput computed using \eqref{eq:e2e_latency} as a function of the block size $B$ for AR-LLM, MDLM, and N-MDLM with different values of the parameter $K$ in \eqref{eq:adaptive_threshold}. 
Recall that  smaller values of $K$ indicate more sparsity.
We also consider a neuromorphic version of AR-LLM, referred to as N-AR-LLM, which applies the same conversion pipeline described in Sec. \ref{sec:conversion} starting from the AR-LLM.
The throughput is normalized  with respect to the AR-LLM baseline so as to highlight the improvements achieved via diffusion and via event-driven processing.
Following the discussion in Sec. \ref{sec:performance_metric}, we focus on the ICMS configuration, but we also show the performance of AR-LLM and MDLM for the OCMS as reference, reflecting a conventional GPU-like implementation for standard models. 

As reported in \cite{arriola2025block}, in the memory-bound regime of OCMS, MDLM is seen to have an increasing throughput as a function of the block size $B$ (dashed line).
In contrast, in the compute-bound regime of ICMS, increasing the block size $B$ does not yield gains for MDLM (solid line).
This is because, in a compute-bound regime, the throughput is limited by the compute rate $L_{\ops}$, and it does not benefit from amortizing memory access via diffusion.

By introducing sparsity, N-MDLM reduces both computation and memory traffic.
This lowers the arithmetic intensity, i.e., the number of operations per memory access, steering the system toward a memory-bound regime.
As a result, in the ICMS configuration, N-MDLM is able to extract performance gains on hardware where plain MDLM shows little improvement.
In particular, the throughput of N-MDLM is seen in Fig. \ref{fig:wmt_throughputb} to increase with block size $B$, eventually saturating when the system enters a compute-bound regime. 

\begin{figure}[!ht]
    \centering
    \includegraphics[width=0.7\linewidth]{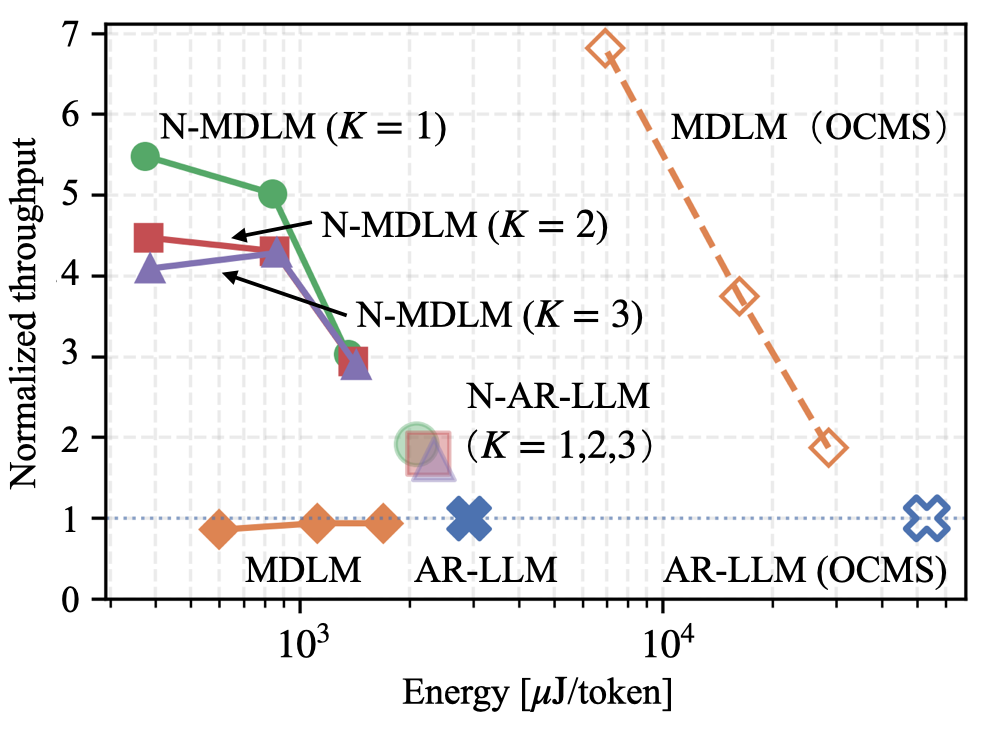}
    \caption{Normalized throughput versus energy per token for AR-LLM, MDLM with $B \in \{2,4,8\}$, N-AR-LLM, and N-MDLM ($B=\{2,4,8\}$), when running on ICMS (solid lines) and OCMS (dashed lines). Throughput is computed using \eqref{eq:e2e_latency}, while energy per token is obtained from \eqref{eq:energy}. The throughput is normalized with respect to the AR-LLM baseline of the corresponding hardware.}
    \label{fig:wmt_throughputb_energy}
    \vspace{-10pt}
\end{figure}

Fig. \ref{fig:wmt_throughputb_energy} illustrates the relationship between throughput, computed via \eqref{eq:latency}, and energy per token, computed via \eqref{eq:energy}, for AR-LLM, N-AR-LLM, MDLM, and N-MDLM with different values of the sparsity-inducing parameter $K$.
The points for MDLM and N-MDLM are obtained by varying the block size $B \in \{2,4,8\}$ (larger values of $B$ correspond to larger energy consumption), while AR-LLM and N-AR-LLM correspond to $B=1$.
Moving from AR-LLM to MDLM, energy consumption decreases due to improved amortization of memory access.
In fact, as shown by the hardware parameters in Table \ref{tab:hw_params}, memory access incurs a significantly higher energy cost than computation.
Therefore, reducing memory traffic is the primary driver of energy efficiency gains in AR-LLM and MDLM. 

N-MDLM further reduces energy consumption, with the lowest energy achieved at the highest sparsity level, i.e., $K=1$.
This trend reflects the effect of event-driven computation, whereby operations are only triggered by spike events. 

\begin{figure}[!ht]
    \centering
    \includegraphics[width=0.7\linewidth]{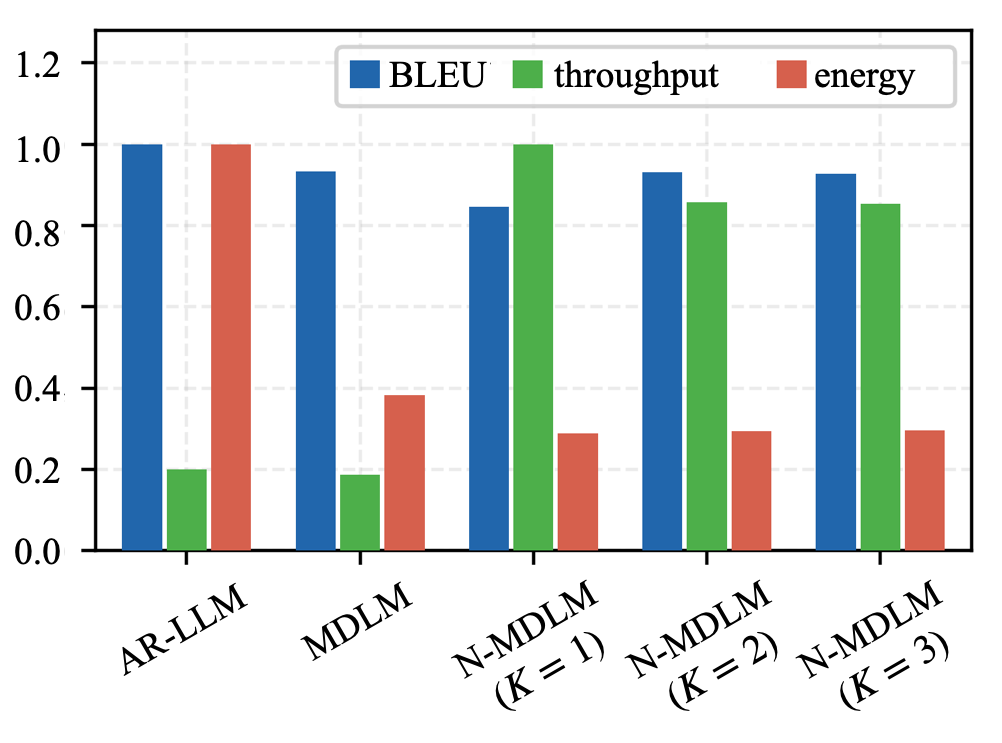}
    \caption{BLEU, energy per token, and throughput across AR-LLM, MDLM ($B{=}4$), and \glsemphpl{nmdlm} ($B{=}4$) in ICMS. 
    Throughput and energy are obtained from \eqref{eq:latency} and \eqref{eq:energy}, respectively.
    All three metrics are independently normalized to the interval $[0,1]$ using their maximum values across configurations.}
    \label{fig:wmt_normalize}
\end{figure}

Finally, Fig.~\ref{fig:wmt_normalize} compares normalized throughput, computed via \eqref{eq:e2e_latency}, energy per token, computed via \eqref{eq:energy}, and BLEU, a measure of accuracy for the given task \cite{bojar-etal-2014-findings}, for AR-LLM, MDLM ($B=4$), and N-MDLM ($B=4$).
The results show that \glsemphpl{nmdlm} outperform AR-LLM and MDLM in terms of both throughput and energy efficiency, with only a slight drop in BLEU, indicating that the efficiency gains are achieved with minimal impact on translation quality.

Overall, the results highlight a trade-off between block size $B$ and sparsity level controlled by $K$.
Augmenting the block size $B$ improves the throughput of a memory-bound system up to the point where it becomes compute bound.
It is in this regime that sparsity becomes effective in improving throughput.
In the considered ICMS, this is achieved for $B=4$ in N-MDLM models with $K=1$ or $K=2$. Augmenting the block size $B$ beyond this point does not yield additional throughput gains. 

With respect to energy, where the cost is dominated by memory access, the system always gains from augmenting $B$ to amortize memory access.

\section{Conclusions}

This paper investigated the integration of masked diffusion language models with neuromorphic computing principles for efficient large language model inference.
Specifically, we proposed \glspl{nmdlm}, which combines block-parallel diffusion decoding with spike-based sparse computation.
While block diffusion increases the number of tokens generated per parameter access, spike-induced sparsity reduces effective synaptic operations by skipping inactive neurons.
We further developed a roofline-inspired analysis that characterizes models in terms of throughput, energy, and accuracy.
Experiments on translation tasks show that the proposed framework improves decoding throughput and energy efficiency while maintaining task performance.
The gains are particularly significant in compute-bound regimes, where plain MDLMs provide limited improvement.
By combining block parallelism with sparsity, \glsemphpl{nmdlm} reduce the computational load and enable throughput gains.
Future work will focus on validation on neuromorphic hardware and on adaptive strategies to jointly optimize $B$ and $K$ under varying system constraints.

\bibliographystyle{IEEEtran}
\bibliography{IEEEabrv, main}

\end{document}